\documentclass[letterpaper]{article} 
\usepackage{aaai20}  
\usepackage{times}  
\usepackage{helvet} 
\usepackage{courier}  
\usepackage[hyphens]{url}  
\usepackage{graphicx} 
\usepackage{multirow}
\usepackage{amsmath}
\usepackage{amssymb}
\usepackage{algorithm}
\usepackage{algorithmic}
\usepackage{float}
\usepackage{latexsym}
\usepackage{color}

\nocopyright

\title{Improving Question Generation with Sentence-level Semantic Matching and Answer Position Inferring}

\author{Xiyao Ma$^1$, Qile Zhu$^1$, Yanlin Zhou$^1$, Xiaolin Li$^2$, Dapeng Wu$^1$ \\
  $^1$NSF Center for Big Learning, University of Florida \\
  $^2$AI Institute, Tongdun Technology \\
  { maxiy@ufl.edu, valder@ufl.edu, zhou.y@ufl.edu,  xiaolin.li@tongdun.net, dpwu@ufl.edu}}

\begin{document}

\maketitle

\begin{abstract}
Taking an answer and its context as input, sequence-to-sequence models have made considerable progress on question generation.
However, we observe that these approaches often generate wrong question words or keywords and copy answer-irrelevant words from the input.
We believe that lacking global question semantics and exploiting answer position-awareness not well are the key root causes.
In this paper, we propose a neural question generation model with two concrete modules: sentence-level semantic matching and answer position inferring.
Further, we enhance the initial state of the decoder by leveraging the answer-aware gated fusion mechanism.
Experimental results demonstrate that our model outperforms the state-of-the-art (SOTA) models on SQuAD and MARCO datasets.
Owing to its generality, our work also improves the existing models significantly.
\end{abstract}

\section{Introduction}

Question Generation (QG), an inverse problem of Question Answering (QA), aims to generate a semantically relevant question given a context and a corresponding answer. 
It has huge potential in education scenario \cite{du2017learning}, dialogue system, and question answering \cite{du2018harvesting}.
A bunch of models using sequence-to-sequence (seq-to-seq) models \cite{sutskever2014sequence} with the attention mechanism \cite{bahdanau2014neural} have been proposed for neural question generation~\cite{zhou2017neural,du2017learning}. 

Enriched linguistic features with Part-Of-Speech (POS) tags, relative position information, and paragraph context are incorporated in the embedding layers \cite{zhou2017neural,kim2018improving,zhao2018paragraph}. 
Copy mechanism \cite{gulcehre2016pointing} is exploited to enhance the output quality of decoders \cite{zhao2018paragraph,sun2018answer}.

\begin{table*}[h]
\centering
\scalebox{1.0}{
\begin{tabular}{l}
    \hline
    \textbf{Sentence}: starting in 1965, donald davies at the national physical laboratory, uk, independently developed \\
    the same \underline{message routing methodology} as developed by baran.\\
    \textbf{Reference}: what did donald davies develop?\\
    \textbf{NQG++}: what is \textcolor{red}{the national physical laboratory}?\\
    \textbf{Pointer-generator}: what did \textcolor{red}{baran} develop?\\
    \textbf{Our model}: what did \textcolor{blue}{donald devies develop} at the national physical laboratory?\\
    \hline
    \textbf{Sentence}:  in 1979 , \underline{the soviet union} deployed its 40th army into afghanistan , attempting to suppress \\
    an islamic rebellion against an allied marxist regime in the afghan civil war.\\
    \textbf{Reference}: who deployed its army into afghanistan in 1979? \\
    \textbf{NQG++}: \textcolor{red}{in what year} did the soviet union invade afghanistan?\\
    \textbf{Pointer-generator}: \textcolor{red}{what} deployed their army into afghanistan?\\
    \textbf{Our model}:  \textcolor{blue}{who} deployed their army into afghanistan? \\
    \hline
\end{tabular}}
\caption{Bad cases of the baselines: the models ask questions with wrong question words and wrong keywords. 
The answers are shown with underline. The text in red indicates the poor performance of existing models, while the text in blue shows the improved performance with our proposed model.}
\label{tab:intro1}
\end{table*}

\begin{table*}[h]
\centering
\scalebox{0.95}{
\begin{tabular}{l}
\hline  
    \textbf{Sentence}:as of 2012 , quality private schools in the united states charged substantial tuition , close to \$ 40,000 \\
    annually for day schools in new york city , and nearly \underline{\$ 50,000} for boarding schools.\\
    \textbf{Reference}: what would a parent have to pay to send their child to a boarding school in 2012?\\
    \textbf{NQG++}:  how much money did \textcolor{red}{quality private schools} in the us have in 2012?\\
    \textbf{Pointer-generator}:  how much money is \textcolor{red}{charged substantial tuition}  for boarding school?\\
    \textbf{Our model}: how much money for \textcolor{blue}{boarding schools} in new york city in 2012? \\
    \hline
    \textbf{Sentence}: during his second year of study at graz , tesla developed a passion for and became very proficient \\
    at billiards , chess and card-playing , sometimes spending \underline{more than 48 hours} in a stretch at a gaming table. \\
    \textbf{Reference}: how long would tesla spend gambling sometimes?\\
    \textbf{NQG++}: how long did \textcolor{red}{the billiards of tesla get} in a stretch?\\
    \textbf{Pointer-generator}: how long did tesla \textcolor{red}{become very proficient} in a stretch at a gaming table ?\\
    \textbf{Our model}: how many hours did tesla \textcolor{blue}{spend in a stretch at a gaming table}?\\
    \hline
\end{tabular}}
\caption{Bad cases of the baselines: the models copy the answer-irrelevant context words from sentences.}
\label{tab:intro2}
\end{table*}

However, checking the questions generated by the strong baseline models NQG++ \cite{zhou2017neural} and Pointer-generator \cite{see2017get} originally solving text summarization, the modern question generation models face the two main issues as follows:
(1) \textit{Wrong keywords and question words:} 
The model may ask questions with wrong keywords and wrong question words, as shown in the examples in Table \ref{tab:intro1}.
(2) \textit{Poor copy mechanism:} The model copies the context words semantically irrelevant to the answer \cite{sun2018answer}, as illustrated in the examples in Table \ref{tab:intro2}. 

Generally, the decoder with parameters $\theta_d$ in seq-to-seq models \cite{zhou2017neural,sun2018answer,zhao2018paragraph} is trained by maximizing the generation probability $p(y_t|y_{<t},z; \theta_d)$ of the reference question word $y_t$, given the previous generated words conditioned on the encoded context $z$. 
However, the decoder may focus on local word semantics while ignoring the global question semantics during generation, resulting in above-mentioned issues. 
Meanwhile, the answer position-aware features are not exploited well by the copy mechanism, resulting in copying answer-irrelevant context words from input.

To alleviate these issues, we claim that learning the sentence-level semantics and answer position-awareness in a Multi-Task Learning (MTL) fashion results in a better performance as shown in Table \ref{tab:intro1} and \ref{tab:intro2}. To do so, we first propose sentence-level semantic matching module for learning global semantics from both the encoder and decoder simultaneously.
Then, answer position inferring module is introduced to enforce the model with the copy mechanism \cite{see2017get} to emphasize the relevant context words with the answer position-awareness.
Furthermore, we propose answer-aware gated fusion mechanism for improved answer-aware sentence vector for decoder.

We further conduct extensive experiments on SQuAD \cite{rajpurkar2016squad} and MS MARCO \cite{nguyen2016ms} dataset to show the superiority of our proposed model.
The experimental results show that our model not only outperforms the SOTA models on main metrics, auxiliary metrics, and human judgments, but also improves different models due to its generality.
Our contributions are three-fold:
\begin{itemize}
    \item We analyze the questions generated by strong baselines and find two issues: wrong keywords and wrong question words and copying answer-irrelevant context words.
    We identify that lacking whole question semantics and expoiting answer position-awareness not weel are the key root causes.
    \item To address the issues, we propose neural question generation model with sentence-level semantic matching, answer position inferring, and gated fusion.
    \item We conduct extensive experiments to demonstrate the superiority of our proposed model for improving question generation performance in terms of the main metrics, auxiliary machine comprehension metrics, and human judgments.
    Besides, our work can improve current models significantly due to its generality.
\end{itemize}

\begin{figure*}[h]
    \centering
    \includegraphics[width=0.9\textwidth]{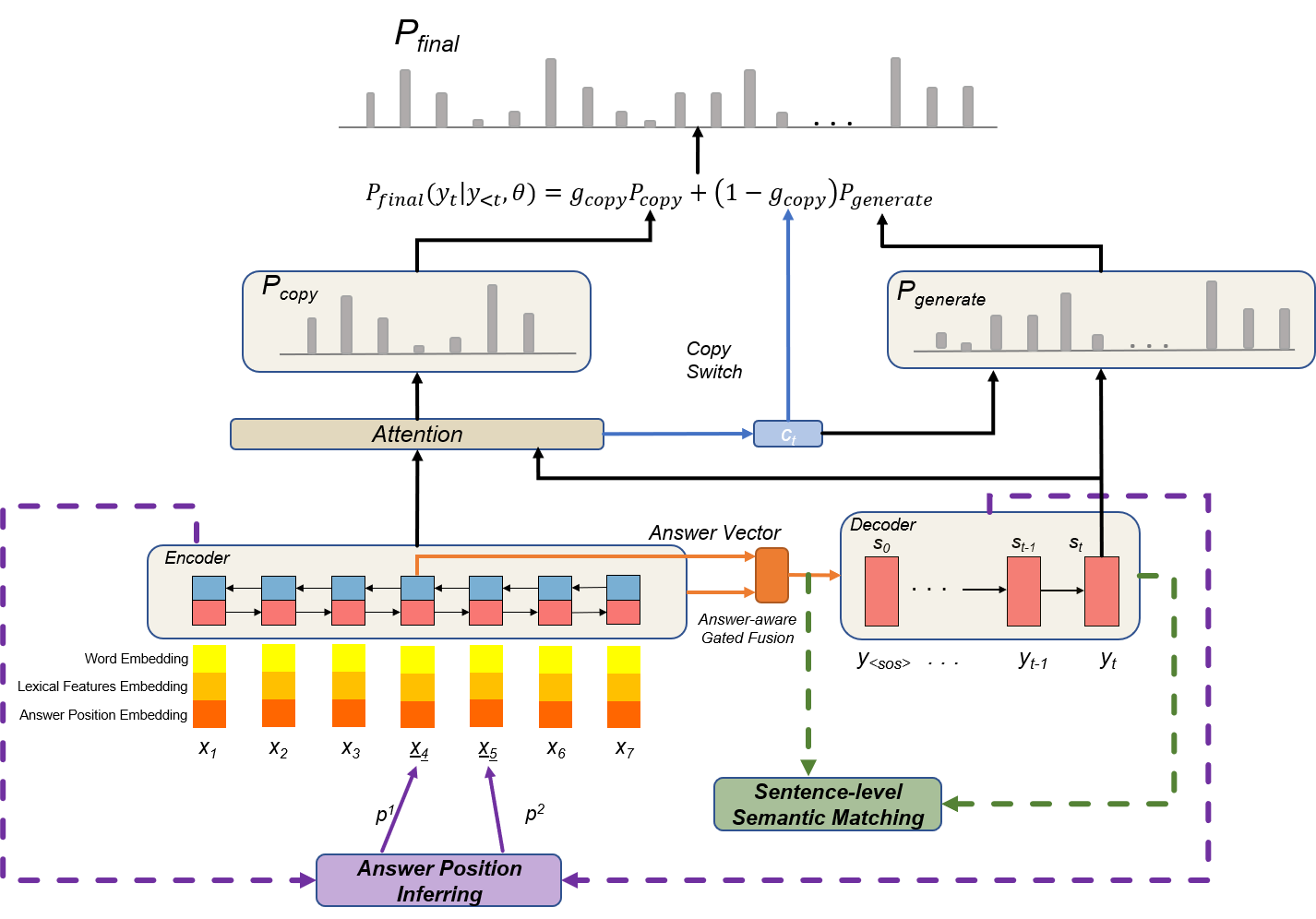}
    \caption{Diagram for neural question generation model with sentence-level semantic matching, answer position inferring, and gated fusion.}
    \label{fig:model}
\end{figure*} 

\section{Proposed Model} \label{modelSection}
In this section, we describe the details of our proposed models, starting with an overview of question generation problem. Then, we illustrate our backbone seq-to-seq model with gated fusion for improved answer-aware sentence vector for generation. 
Finally, we illustrate sentence-level semantic matching and answer position inferring to alleviate the issues we discussed in the previous section.

\subsection{Problem Formulation}
In a question generation problem, a sentence $X=\{x_t\}^M_{t=1}$ containing an answe $A$, a contiguous span of the sentence, is given to generate a question $Y=\{y_t\}^N_{t=1}$ matching with the sentence $X$ and the answer $A$ semantically.

\subsection{Seq-to-seq model with Answer-aware Gated Fusion}

\paragraph{Encoder:} 
Following the baseline model \cite{zhou2017neural}, we use an attention-based seq-to-seq model with the same enriched semantic and lexical features (i.e., NER features \cite{sang2003introduction}, POS tag \cite{brill1992simple}, case, and answer position features) as input $x_i \in \mathbb{R}^{(d_w+d_n + d_p + d_c + d_{ap})}$ in the embedding layer.

With a bi-directional LSTM \cite{hochreiter1997long} as the encoder, the sentence representation, a sequence of D-dim hidden state $H=[h_1,h_2,...,h_m] \in \mathbb{R}^{M*D}$, is produced by concatenating a forward hidden state and a backward hidden state given the input sentence $X=[x_1,x_2,...,x_m]$:

\begin{align} 
    & h_i = [\overrightarrow{h_i}, \overleftarrow{h_i}], \\
    & \overrightarrow{h_i} = 
    LSTM_{Enc}(x_i, \overrightarrow{h_{i-1}}), \\
    & \overleftarrow{h_i} = LSTM_{Enc}(x_i,\overleftarrow{h_{i+1}} )
\end{align}

\paragraph{Answer-aware Gated Fusion:} 
Instead of passing the last hidden state $h_m$ of the encoder to the decoder as the initial hidden state, we propose gated fusion to provide an improved answer-aware sentence vector $z$ for the decoder. 

Similar to the gates in LSTM, we use two information flow gates computed by $Sigmoid$ functions to control the information flow of sentence vector and answer vector:
\begin{align}
    g_m &= \sigma (W_m^T*[h_m, h_a] + b_m) , \\
    g_a &= \sigma (W_a^T*[h_m, h_a] + b_a), \\
    z &= g_m \cdot h_m + g_a \cdot h_a
\end{align}
where $W_m$, $W_a$, $b_m$, and $b_a$ are trainable weights and biases.
We take the hidden state at the answer starting position as the answer vector $h_{a} \in \mathbb{R}^{D}$ since it encodes the whole answer semantics with the bi-directional LSTM.

\paragraph{Decoder:}
Taking the encoder hidden states $H=[h_1,h_2,...,h_m] \in \mathbb{R}^{M*D}$ as the context and the improved answer-aware sentence vector $z$ as the initial hidden state $s_1$, an one layer uni-directional LSTM updates its current hidden state $s_t$ with the previous decoded word as the input $w_t$:
\begin{align} 
    & s_t = LSTM_{Dec}([w_t; c_{t-1}], s_{t-1})
\end{align}

Meanwhile, the attention mechanism \cite{bahdanau2014neural} is exploited by attending the current decoder state $s_t$ to the encoder context $H=[h_1,h_2,...,h_m]$. 
The context vector $c_t$ is computed with normalized attention vector $\alpha_t$ by weighted-sum:
\begin{align}
    & e_t =v_{a}^{T} \tanh (W s_t + U H)\\
    & \alpha_t = Softmax(e_{t}), \label{softmax} \\
    & c_t = H^T \alpha_t
\end{align}

Question word $y_t$ is generated from vocabulary $V$ with $Softmax$ function:
\begin{align}
    & p_{generate}(y_t) = Softmax(f(s_t, c_t))
\end{align}
where $f$ is realized by a two-layer feed-forward network.

\paragraph{Copy Mechanism / Pointer-generator:} Copy Mechanism \cite{gulcehre2016pointing} and Pointer-generator network \cite{see2017get} are introduced to enable the model to generate words from the vocabulary $\mathbb{V}$ with size $|\mathbb{V}|$ or copy words from the input sentence $\mathbb{X}$ with size $|\mathbb{X}|$ by taking the $i^{th}$ word with the highest attention weight $\alpha_{t,i}$ computed in Equation ~\ref{softmax}.

Generally, when generating the question word $y_t$, a copy switch $g_{copy}$ is computed to decide whether the generated word is generated from vocab or copied from source sentence, given the current decoder hidden state $s_t$ and context vector $c_t$:
\begin{align}
g_{copy}=\sigma(W^cs_t + U^cc_t + b^c)
\end{align}
where $W^c$, $U^c$, and $b^c$ are learnable weights and biases.

The final word distribution is obtained by combining the probability of generate mode and the probability of copy mode:
\begin{equation}
\begin{split}
    p_{\text {final}}\left(y_{t} | y_{<t}; \theta_{s2s}\right)&=g_{copy} p_{\text {copy}}(y_t, \theta_1)\\
    &+\left(1-g_{\text {copy}}\right) p_{\text {generate}}(y_t, \theta_2)
\end{split}
\end{equation}
where $\theta$, $\theta_1$, and $\theta_2$ are the parameters of neural network.

We use the negative log likelihood for the seq-to-seq loss:
\begin{align}
    L\left(\theta_{s2s}\right)=-\frac{1}{N} \sum_{i}^{N} \log p_{\text {final}}\left(y_{t} | y_{<t}; \theta_{s2s}\right)
\end{align}
where $\theta_{s2s}$ is the parameters of the seq-to-seq model, and N is the number of data in the train dataset.

\subsection{Sentence-level Semantic Matching}
Existing models, especially the decoders, generate question words given the generated and partial question words without considering the global whole question semantic, prone to wrong question words or keywords. 
Meanwhile, we observe that there exist different reference questions targeting the different answers in the same sentence in SQuAD and MARCO datasets.
For example,  we have $<sentence, answer1, question1>$ and $<sentence, answer2, question2>$.
However, the baseline model is prone to generating generic questions in this case.
To overcome this problem, we propose the sentence-level semantic matching module to learn the sentence-level semantics from both the encoder and decoder sides in a multi-task learning way.
\begin{figure}[h]
    \centering
    \includegraphics[width=0.4\textwidth]{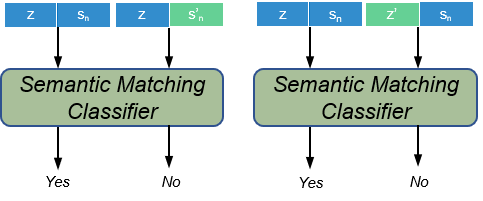}
    \caption{Pipeline of Sentence-level Semantic Matching}
    \label{fig:sm}
\end{figure} 

Generally, we have the improved answer-aware sentence vector $z$ obtained by our gated fusion. 
Regarding the decoder, a uni-directional LSTM, as an encoder for question, we take the last hidden state $s_n$ as the question vector. 

Then, as illustrated in Figure \ref{fig:sm}, we train two classifiers to distinguish the non-semantic-matching sentence-question pairs from semantic-matching pairs given their sentence-level representation vector and labels $[z^i, s_n^i, \ell_{i}]$, which are uniformly sampled from the current batch data:
\begin{align}
    p_{sm} = Softmax(W_c[z^i, s_n^i]+ b_c)
\end{align}
where $[z^i, s_n^i]$ is the concatenation of the sentence vector $s_n^i$ and the question vector $s_n^i$.

We jointly train the classifiers with seq-to-seq model in a multi-task learning fashion by minimizing the cross entropy loss:
\begin{align}
    L_{sm}(\theta_{s2s}, \theta_{sm}) =-\mathbb{E}_{(z^i, s_n^i, \ell_{i})}\log p_{sm}(\ell_{i} | z^i, s_n^i; \theta_{sm})
\end{align}
where $\theta_{sm}$ is the parameters of the two classifiers, and $p_{}sm$ is the output probability of the classifiers. 

\subsection{Answer Position Inferring}
Another issue of the baseline model is that it copies the answer-irrelevant words from the input sentence.
One potential reason is that the model does not learn the answer position features well, and the attention matrix is not signified by the context words relevant to the answer.
To address the issue, we leverage answer position inferring module to steer the model to learn the answer position-awareness, still in a Multi-Task Learning fashion.
\begin{figure}[h]
    \centering
    \includegraphics[width=0.35\textwidth]{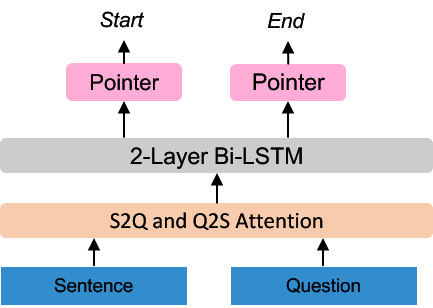}
    \caption{Framework of Answer Position Inferring}
    \label{fig:pointer}
\end{figure} 
We borrow the bi-directional Attention Flow network and output layer from BiDAF model \cite{seo2016bidirectional} to infer the answer position as shown in Figure \ref{fig:pointer}, taking the sentence representation $H \in \mathbb{R}^{M*D}$ and question representation $S \in \mathbb{R}^{N*D}$ from the encoder and the decoder as inputs.

Formally, we take Sentence-to-Question (S2Q) attention and Question-to-Sentence (Q2S) attention to emphasize the mutual semantic relevance between each sentence word and each question word, and we obtain the question-aware sentence representation $\tilde{{H}}$ and the sentence-aware question representation $\tilde{{S}}$ by using similar attention mechanism to Equation \ref{softmax}:
\begin{align}
    \tilde{{H}} = attn(H, S), \\
    \tilde{{S}} = attn(S, H)
\end{align}

\begin{table*}[t]
\centering
\caption{Comparison of models performances in terms of the main metrics on SQuAD and MARCO dataset}
\scalebox{0.64}{
\begin{tabular}{l|cccccc|cccccc}
\hline
 & \multicolumn{6}{c|}{SQuAD} & \multicolumn{6}{c}{MARCO} \\ \hline
Models & BLEU-1 & BLEU-2 & BLEU-3 & BLEU-4 & METEOR & ROUGE-L & BLEU-1 & BLEU-2 & BLEU-3 &  BLEU-4 & METEOR & ROUGE-L \\ 
 \hline
NQG++ \cite{zhou2017neural} & 42.13 & 25.98 & 18.24 & 13.29 & 17.59 & 40.75 & 46.62 & 32.67 & 22.98 & 16.13 & 20.22 & 46.35 \\ 
Pointer-generator \cite{see2017get} & 42.43 & 26.75 & 18.99 & 14.33 & 18.77 & 43.19 & 47.10 & 34.26 & 24.87 & 17.95 & 22.34 & 47.69 \\ 
Answer-focused \cite{sun2018answer} & 43.02 & 28.14 & 20.51 & 15.64 & - & - & - & - & - & - & - & -\\ 
Gated Self-attention \cite{zhao2018paragraph} & 44.51 & 29.07 & 21.06 & 15.82 & 19.67 & 44.24 & - & - & - & - & - & -\\ \hline
Model with Sentence-level Semantic Matching & 43.67 & 28.53 & 20.59 & 15.66 & 19.23 & 43.86 & 48.97 & 35.84 & 26.31 & 19.79 & 23.83 & 48.93\\ 
Model with Answer Position Inferring & 43.88 & 28.55 & 28.87 & 15.77 & 19.55 & 43.98 & 49.73 & 36.77 & 26.46 & 20.14 & 24.22 & 49.33\\ 
Combined Model & \textbf{44.71} & \textbf{29.89} & \textbf{21.77} & \textbf{16.32} & \textbf{20.84} & \textbf{44.79} & \textbf{50.33} & \textbf{37.10} & \textbf{27.23} & \textbf{20.46} & \textbf{24.69} & \textbf{49.89}\\ \hline
\end{tabular}
}
\label{tab:mainMetrics}
\end{table*}

Then, two two-layer bidirectional LSTMs are used to capture the interactions among the sentence words conditioned on the question \cite{seo2016bidirectional}. 
The answer starting index and end index are predicted by the output layer with $Softmax$ function:
\begin{align}
    M_1 &= LSTM(f(H, \tilde{{H}}, \tilde{{S}})), \\
    M_2 &= LSTM(M_1), \\
    {p}^{1}&=\operatorname{Softmax}\left({W}_{\left({p}^{1}\right)}^{\top}[\tilde{{H}}, M_1]\right), \\
    {p}^{2}&=\operatorname{Softmax}\left({W}_{\left({p}^{2}\right)}^{\top}[\tilde{{H}}, M_2]\right)
\end{align}
where $W_{p^1}$ and $W_{p^2}$ are trainable weights, and $f$ function is a trainable multi-layer perception (MLP) network.

We compute the loss with the negative log likelihood of the ground truth answer starting index $y_i^1$ and ending index $y_i^2$ with the predicted distribution:
\begin{align}
    L(\theta_{ap}, \theta_{s2s})=-\mathbb{E}_{(y_i^1, y_i^2)}\log \left({p}_{y_{i}^{1}}^{1}\right)+\log \left({p}_{y_{i}^{2}}^{2}\right)
\end{align}
where $\theta_{ap}$ is the parameters to be updated of the answer position inferring module, and $y_i^1, y_i^2$ are the ground truth answer position labels.

To jointly train the generation model with the proposed modules in a multi-task Learning approach, we minimize the total loss during the training:
\begin{align}
    L(\theta_{all}) = L(\theta_{s2s}) + \alpha*L(\theta_{sm}, \theta_{s2s}) + \beta * L(\theta_{ap}, \theta_{s2s})
\end{align}
where $\alpha$ and $\beta$ control the magnitude of the sentence-level semantic matching loss and the answer position inferring loss, and $\theta_{all}$ are all the parameters of our model.
By minimizing the above loss function, our model is expected to discover and utilize the sentence-level and answer position-aware semantics of the question and sentence.

\section{Experiments and Results}
In this section, we conduct extensive experiments on the SQuAD and MS MARCO dataset, demonstrating the superiority of our proposed model compared with existing approaches.

\subsection{Experiment Settings}

\paragraph{Dataset} SQuAD V1.1 dataset contains 536 Wikipedia articles and more than 100K questions posed about the articles \cite{rajpurkar2016squad}. 
The answer is also given with corresponding questions as the sub-span of the sentence. 
Following the baseline \cite{zhou2017neural}, we use the training dataset (86635) to train our model, and we split the dev dataset into dev (8965) and test dataset (8964) with a ratio of 50\%-50\% for evaluation.

MS MARCO contains more than one million queries along with answers either generated by human or selected from passages \cite{nguyen2016ms}. 
We select a subset of MS MARCO, where the answers are sub-spans of the passages. 
We split them into train set (86039), dev set (9480), and test set (7921) for model training and evaluation purpose.

We report automatic evaluation with BLEU-1, BLEU-2, BLEU-3, BLEU-4 \cite{papineni2002bleu}, METEOR \cite{denkowski2014meteor}, and ROUGE-L \cite{lin2004rouge} as the main metrics.

\paragraph{Baselines} In the experiments, we have several baselines for comparisons:
\begin{itemize}
    \item NQG++ \cite{zhou2017neural}: It is a baseline for Neural Question Generation task. 
    It uses enriched semantic and lexical features in the encoder embedding layer of the seq-to-seq model. 
    Attention mechanism and copy mechanismare also used.
    \item Feature-enriched Pointer-generator \cite{see2017get}: It is a seq-to-seq model with attention mechanism and copy mechanism. The copy mechanism is realized differently from NQG++.
    We add enriched features used in NQG++ in the embedding layer.
    \item Answer-focused \cite{sun2018answer}: It is a SOTA model on QG that uses an additional vocabulary for question word generation with relative answer position information instead of BIO used in NQG++.
    \item Gated Self-attention \cite{zhao2018paragraph}. It is also a SOTA model on QG that leverages paragraph as input with gated self-attention above RNN in the encoder. Meanwhile, an improved maxout pointer is introduced. 
\end{itemize}

\subsection{Results and Analysis}
\paragraph{Main Metrics} We report the main metrics of different models on SQuAD and MS MARCO dataset in Table \ref{tab:mainMetrics}.

Answer-focused model \cite{sun2018answer} improves the performance by using separate vocabulary for question word generation along with answer relative position.
The Gated Self-Attention model \cite{zhao2018paragraph} emphasizes the intra-attention among the sentence with improved maxout pointer.

Different from the models above, our work aims to improve the model by learning the sentence-level semantic-matching features on both the encoder and decoder sides. 
The result shows that our model outperforms the two SOTA models on the main metrics.

\begin{table}
\centering
\caption{Machine Comprehension Performance in terms of Exact Match (EM) and F1on SQUAD dataset}
\scalebox{0.82}{\begin{tabular}{l|lr}
\hline
Questions     & EM (\%) & F1 (\%) \\ 
\hline
Reference Questions        & 49.68   &  65.97  \\
\hline
NQG++ \cite{zhou2017neural}  &  35.26  &  50.88  \\
Pointer-generator \cite{see2017get}  &  38.89  &  54.06  \\
\hline
Our model  &  \textbf{42.70}  &  \textbf{57.68}  \\
\hline
\end{tabular}}
\label{tab:MCmetrics}
\end{table}

\paragraph{Auxiliary Metrics} Although the main metrics can reflect the similarity between the generated question and the references, it has its limits on reflecting the semantics of generated question \cite{xu2018lsdscc}.

Alternatively, considering that machine comprehension takes the article and the corresponding question as the input to find the answer in the passages, we adopt the machine comprehension metrics \cite{rajpurkar2016squad} to evaluate the quality of the questions generated by different models \cite{wang2017gated}.

We show the performances of BiDAF \cite{seo2016bidirectional} pre-trained by AllenNLP \cite{Gardner2017AllenNLP} in terms of Exact Match (EM) and F1 metrics on reference questions, questions generated by baseline, and questions generated by our model in Table \ref{tab:MCmetrics}.

Our model outperforms NQG++ and Pointer-generator on EM and F1 significantly, since our model generates more answer-relevant questions by discovering sentence-level semantics and answer position features.

\paragraph{Sentence-level Semantic Matching Analysis} To analyze the quality of our model on generating the right question words and keywords, we randomly sample 200 questions generated by NQG++, Pointer-generator, and our model, respectively.
Generally, the generated question is claimed to have the right question words if it has the same question words to the reference question.
For example, we have a generated question "what place ..." and a reference question "where ...", and we claim that the model generate a question with the right question words.
In addition, we choose the words with most semantics importance as the keywords, which indicate the sentence topic and content.
We report the number of the questions with right question words and keywords by different models in Table \ref{tab:SMA}.

\begin{table}[h]
\caption{Question words and keywords generation performance by different models on SQuAD dataset}
\scalebox{0.6}{
\begin{tabular}{l|cc}
\hline
Models & \# right question words & \# right keywords \\
\hline
NQG++ \cite{zhou2017neural} &  134 & 143\\
Pointer-generator \cite{see2017get} &  140 & 148\\
 \hline
Model with Sentence-level Semantic Matching & \textbf{150} & \textbf{156}\\
\hline
\end{tabular}}
\label{tab:SMA}
\end{table}

The main reason that our model outperforms the existing model is that learning the sentence-level semantics helps to capture the key semantics and results in better performance on generating the semantic-matching keywords.

\paragraph{Answer Position Inferring Analysis} We also conduct the similar experiment on evaluating the copy mechanisms in different models in terms of precision and recall used in \cite{sun2018answer}.
Given one generated question $\text{G}$ and reference question $\text{R}$, we definite precision and recall as:

\begin{align}
    \text{Precision} =& \frac{\sum_{i}^{N} \text{\# OOV words in both $G_i$ and $R_i$}}{\sum_{i}^{N} \text{\# OOV words in $G_i$}} \\
    \text{Recall} =& \frac{\sum_{i}^{N} \text{\# OOV words in both $G_i$ and $R_i$}}{\sum_{i}^{N} \text{\# OOV words in $R_i$}}
\end{align}

\begin{table}[h]
\caption{Copy mechanism performance by different models}
\scalebox{0.8}{
\begin{tabular}{l|cc}
\hline
Models & Precision & Recall\\
\hline
NQG++ \cite{zhou2017neural} & 46.28\% & 32.13\% \\
Pointer-generator \cite{see2017get} & 47.21\% & 38.38\% \\
 \hline
Model with Answer Position Inferring & \textbf{48.35\%} & \textbf{40.27\%}\\
\hline
\end{tabular}}
\label{tab:copy}
\end{table}

As reported in Table \ref{tab:copy}, the improvement of Precision and Recall indicates that answer position inferring can help copy OOV words from the input sentence. 

\paragraph{Model Generality} To show the effectiveness and generality of our work, we evaluate the validness of our work by applying it to current representative models without revising the models.
As shown in Table \ref{tab:general}, our work can improve existing models by more than 2\% on QG tasks due to its effectiveness and generality. 
\begin{table}[h]
\caption{Performance Improvement on existing models on SQuAD dataset}
\scalebox{0.72}{
\begin{tabular}{l|c}
\hline
Models & BLEU-4 \\
\hline
NQG++ \cite{zhou2017neural} & 13.29 \\
NQG++ \cite{zhou2017neural} + our work &  \textbf{14.97}\\
\hline
Pointer-generator model \cite{see2017get} &  14.33\\
Pointer-generator model \cite{see2017get} + our work &  \textbf{16.32} \\
\hline
\end{tabular}}
\label{tab:general}
\end{table}

\subsection{Human Evaluation}
We also conduct human evaluation to examine the quality of the questions generated by the models and reference questions by scoring them on a scale of 1 to 5 in terms of semantics matching, fluency, and syntactically correctness.
As reported in Table \ref{tab:human}, our model generates questions with higher scores on the three metrics than the two baseline models,
indicating the superiority of our proposed model by utilizing the sentence-level semantics and answer position-awareness.

\begin{table}[h]
\caption{Human evaluation on questions generated by the models}
\scalebox{0.7}{
\begin{tabular}{l|ccc}
\hline
Models            & Semantic Matching & Fluency & Syntactically Correctness \\ \hline
NQG++             &    1.88     &     2.70     &    3.26     \\ \hline
Pointer-generator &     2.34    &      3.2    &      3.5   \\ \hline
Our model         &     \textbf{2.87}    &     \textbf{3.46}     &    \textbf{3.89}     \\ \hline
\end{tabular}}
\label{tab:human}
\end{table}

\begin{table*}[!t]
\centering
\scalebox{1.0}{
\begin{tabular}{l}
    \hline
    \textbf{Sentence}: another example was the insignificance of \underline{the ministry of war} compared with native chinese \\
    dynasties , as the real military authority in yuan times resided in the privy council.\\
    \textbf{Reference}: who had no real military power during the yuan?\\
    \textbf{NQG++}: the insignificance of \textcolor{red}{what war} was compared to native chinese dynasties?\\
    \textbf{Pointer-generator}: what was the insignificance?\\
    \textbf{Our model}: what was insignificance compared with native Chinese dynasties?\\
    \hline
    \textbf{Sentence}: another example was the insignificance of the ministry of war compared with native chinese \\
    dynasties , as the real military authority in yuan times resided in \underline{the privy council}.\\
    \textbf{Reference}: who had military control during the yuan?\\
    \textbf{NQG++}: what did the chinese dynasties \textcolor{red}{call the insignificance of the ministry of war}? \\
    \textbf{Pointer-generator}: in where \textcolor{red}{the insignificance as} the real military authority in yuan times? \\
    \textbf{Our model}: \textcolor{blue}{the real military authority} in yuan times \textcolor{blue}{resided} where?\\
    \hline
\end{tabular}}
\caption{Examples of questions with asking about the right keywords generated by our model.}
\label{tab:casestudy1}
\end{table*}

\subsection{Case Study}
In this section, we present some examples of questions generated by our model. 

Furthermore, we present a pair of examples, which have the same input sentence in Table \ref{tab:casestudy1}.
Different from that NQG++ generate similar and non-semantic-matching questions, our model can ask different and more semantic-matching questions than baselines, targeting the different answers.

\subsection{Implementation Details} 
Followed NQG++ \cite{zhou2017neural}, we conduct our experiments on the preprocessed data provided by \cite{zhou2017neural}. We use 1 layer LSTM as the RNN cell for both the encoder and the decoder, and a bidirectional LSTM is used for the encoder. 
The hidden size of the encoder and decoder are 512. 
We use a 300-dimension pre-trained Glove vector as the word embedding \cite{pennington2014glove}. 
As same as NQG++ \cite{zhou2017neural}, the dimensions of lexical features and answer position are 16. 
We use Adam \cite{kingma2014adam} Optimizer for model training with an initial learning rate as 0.001, and we halve it when the validation score does not improve. During the training of  Sentence-level Semantic Matching module, we sample the negative sentences and questions from nearby data samples in the same batch, due to the preprocessed data \cite{zhou2017neural} lacking of the information about which data samples are from the same passage.
We compute our total loss function with $\alpha$ of 1 and $\beta$ of 2.
Models are trained for 20 epochs with mini-batch of size 32. 
We choose model achieving the best performance on the dev dataset.

\section{Related Work}
Question generation tasks can be categorized into two classes: one is the rule-based method, meaning manually design lexical rules or templates to convert context into questions without deep understanding on the context semantic \cite{mazidi2014linguistic,labutov2015deep}.
The other one is neural network based methods, which adopt seq-to-seq \cite{sutskever2014sequence} or an encoder-decoder \cite{cho2014learning} framework to generate question words from scratches \cite{du2017learning,zhou2017neural}.
Our work focuses on the second category.

\cite{du2017learning} firstly proposes to generate question with a seq-to-seq model given a context automatically. 
However, the model does not take the answer into consideration. 
Then \cite{zhou2017neural} proposes to use a feature-enriched encoder to encode the input sentence by concatenating word embedding with lexical features as the encoder input, and answer position are involved in informing the model where the answer is.
It is shown that it brings considerable improvements to the model. 
With the success of reinforcement learning, \cite{yuan2017machine} propose to combine supervised learning and reinforcement learning together for question generation by using policy gradient after training the model in supervised learning way. The reward term in the policy gradient loss function can be perplexity and the BLEU scores \cite{papineni2002bleu}.
To tackle the issue that question words do not match with the answer type, \cite{sun2018answer} introduce a vocabulary only to generate question words. 
\cite{zhao2018paragraph} propose to use paragraph as the input for providing more semantic information with an improved maxout pointer for copying words from the input.

Different from existing methods focusing on utilizing more informative features and improving the copy mechanism, we point out that incapability of capturing sentence-level semantics and exploiting answer-aware features are the main reasons, and we alleviate the problem by proposing two modules which can be integrated with any base models named sentence-level semantic matching and answer position inferring in Multi-Task Learning fashion.

\section{Conclusion}
In this paper, we observe two issues with the widely used baseline model on question generation. 
We point out the root cause is that existing models neither consider the whole question semantics nor exploit the answer position-aware features well.
To address the issue, we propose the neural question generation model with sentence-level semantic matching, answer position inferring, and gated fusion.
Extensive experimental results show that our work improves existing models significantly and outperforms the SOTA models on SQuAD and MARCO datasets.

\section{Acknowledgement}
This research was supported by CBL industry and agency members and by the IUCRC Program of the National Science Foundation under Grant No. CNS-1747783.

\bibliographystyle{aaai}
\bibliography{nqg}

\bigskip

\end{document}